\documentclass{article} 
\usepackage{graphicx} 
\usepackage[
backend=biber,
style=numeric,
]{biblatex}
\usepackage{subcaption} 
\usepackage{tikz}
\usetikzlibrary{positioning}
\usepackage{booktabs}
\usepackage{placeins}
\usepackage{amsmath}
\usepackage{hyperref}

\usepackage{algorithm}
\usepackage{algpseudocode}
\usepackage{xcolor}

\title{Reading the unreadable: Creating a dataset of 19th century English newspapers using image-to-text language models}
\author{Jonathan Bourne}
\date{June 2024}

\begin{document}

\maketitle

\begin{abstract}

Oscar Wilde said, ``The difference between literature and journalism is that journalism is unreadable, and literature is not read". Unfortunately, The digitally archived journalism of Oscar Wilde's 19th century often has no or poor quality Optical Character Recognition (OCR), reducing the accessibility of these archives and making them unreadable both figuratively and literally. This paper helps address the issue by performing OCR on "The Nineteenth Century Serials Edition" (NCSE), an 84k-page collection of 19th-century English newspapers and periodicals, using Pixtral 12B, a pre-trained image-to-text language model. The OCR capability of Pixtral was compared to 4 other OCR approaches, achieving a median character error rate of 1\%, 5x lower than the next best model. The resulting NCSE v2.0 dataset features improved article identification, high-quality OCR, and text classified into four types and seventeen topics. The dataset contains 1.4 million entries, and 321 million words. Example use cases demonstrate analysis of topic similarity, readability, and event tracking. NCSE v2.0 is freely available to encourage historical and sociological research. As a result, 21st-century readers can now share Oscar Wilde's disappointment with 19th-century journalistic standards, reading the unreadable from the comfort of their own computers.

\end{abstract}

\section{Introduction}

Britain went through an enormous period of change in the 19th century. Inventions such as the steam train, telegraph and anaesthetic were life-changing for millions. It was also a time of immense societal change as Britain began the transition from what modern eyes would consider a sort of industrial feudalism into a liberal democracy. For example, slavery was banned \cite{uk-government_slavery_1833}, The proportion of the population eligible to vote increased dramatically \cite{johnson_history_2013, phillips_great_1995}, married women were allowed to own property \cite{uk-government_married_1882}, and children under 10 were banned from working in Mines \cite{ashley-cooper_act_1842}. Throughout this period, improvements in literacy \cite{stone_literacy_1969} and printing technology meant that the changes occurring throughout the country were reported on by journalists and discussed in the many local and national newspapers and periodicals. Indeed, so expansive was the change in literacy and the popularity of the press that even journalism went through substantial changes. The traditional elite-centred journalism gave way to the ``New Journalism" \cite{hampton_new_2008, twycross_rise_2024} of the late 1800's. This ``New Journalism" focused more on the mass market audience, but was described as ``feather brained"  \cite{arnold_up_1887} for its perceived low-brow style and possibly being the reason for the quip \cite{wilde_critic_1891} quoted in the abstract of this paper.

Many 19th-century newspapers and periodicals have survived into the 21st century in original form or as microfilm images. Such archival newspapers are invaluable to modern scholars to understand what happened during the 19th century and how it was perceived at the time. Such understanding can help guide modern society when considering major changes now, dealing with polarizing issues, or morally ambiguous subjects. To make use of these archival resources requires searchable digital databases. However, whilst there has been a substantial effort to digitize historical archives \cite{terras_rise_2011}, the process has proved challenging \cite{smith_research_2018, chiron_impact_2017}. 

One such historical Archive is the Nineteenth-century Serial Edition (NCSE) \cite{brake_nineteenth-century_2008}. A collection of 6 periodicals spanning almost the entire nineteenth century, including over 4000 issues and approximately 84,000 pages of data. The original attempt at Optical Character Recognition (OCR) was not successful, and recently, there have been attempts to recover the original text using Transformer Language Model (LM) based post-OCR correction techniques \cite{bourne_clocr-c_2024, bourne_scrambled_2024-1}. However, these approaches were only partially successful. 

Recently the release of easy-to-use image-to-text models \cite{agrawal_pixtral_2024, wei_general_2024, beyer_paligemma_2024} and deep learning approaches to page layout analysis \cite{auer_docling_2024, dell_american_2024, almutairi_newspaper_2024}, suggest that the time is ripe for a more robust open source approach to historical archive digitization. These advances allow for the creation of a simple two-part process for extracting formatted text from historical newspapers, which is easily adaptable to new languages.

\subsection{Related work}

Due to the importance of the digitisation of historical archives and the challenges faced, there has been substantial research in the area of OCR and automated page layout analysis \cite{girdhar_digitizing_2024}. However, whilst there are many recent OCR and Layout models \cite{wei_general_2024, da_vision_2023, zhang_vsr_2021, li_dit_2022, huang_layoutlmv3_2022}, most of them are designed for extracting text from benchmarks such as scientific articles \cite{zhong_publaynet_2019}, receipts \cite{park_cord_2019}, or forms \cite{jaume_funsd_2019}, not historical documents. This is an important distinction as historical documents have dense text-heavy layouts \cite{sven_page_2022}, resulting in poor layout parsing and text extraction. 

Post-OCR correction using text-to-text autoregressive Language Models has recently gained traction as an approach to dealing with the poor OCR of historical documents \cite{boros_post-correction_2024, thomas_leveraging_2024, bourne_clocr-c_2024, bourne_scrambled_2024-1, kanerva_ocr_2025}. This approach is based on the ability of the transformer architecture to maintain long-range dependencies across the text \cite{vaswani_attention_2017}, its ability to have a representation of language \cite{kallini_mission_2024}, and its skill at infilling \cite{devlin_bert_2019} and crucially generative power. Whilst this has had some success, seeing reductions of over 50\% in error rates \cite{thomas_leveraging_2024, bourne_scrambled_2024-1}, this approach is inherently limited as it uses the corrupted output of the OCR process as opposed to the original images. 

The increasing prevalence and ability of multi-modal language models \cite{agrawal_pixtral_2024, team_gemini_2024, openai_gpt-4_2024, qwen_qwen25_2025, dubey_llama_2024}, allows the input of images and the return of text. Many of these models have shown the ability to detect and read text; however, as yet, their ability to cope with the challenges posed by historical documents is under-explored. The value they provide is the ability to respond to prompts made of interleaved text and images and then draw on both the spatial context within the image and the broader knowledge context embedded in their parameters to provide relevant responses. 

Such an ability means image-to-text models are not restricted to simply extracting text but can also be used to embed the entire image and query the embedding \cite{faysse_colpali_2024}; in many ways, this allows issues related to layout and OCR to be entirely circumvented. However, OCR reduces the image to its core information, that is, the words on the page, which are easily verifiable, can be searched and read by a human and require very little storage space. This is not to say that there is no place for image embedding in the digitisation of historical documents, but the vast majority of the value they contain is not related to the image and layout but to the writing itself. As such, this paper focuses on OCR.

\begin{figure}
    \centering
    \includegraphics[width=\linewidth]{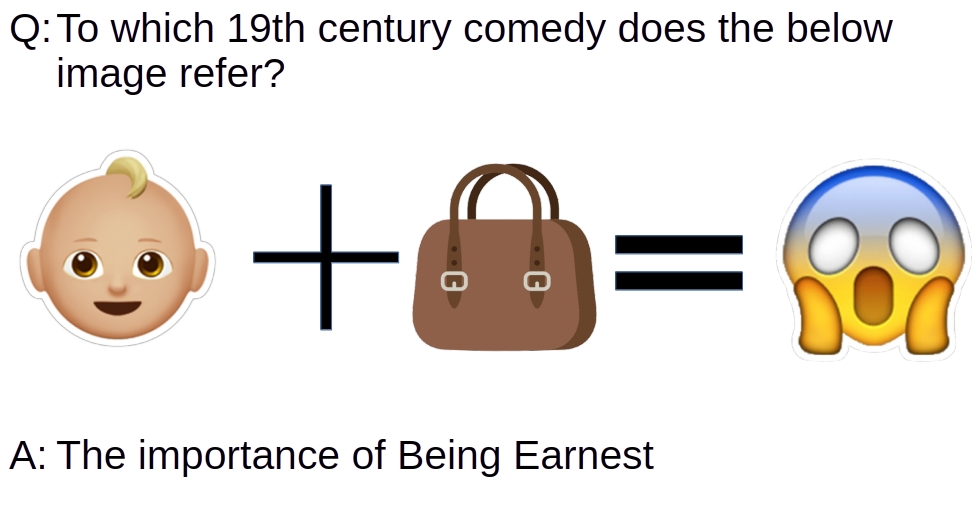}
    \caption{Multi-modal Language models can interleave text and images, allowing them to infer image based cultural references from textual cues. This is essential for effective OCR. }
    \label{fig:importance_earnest}
\end{figure}

Figure \ref{fig:importance_earnest} provides an example of how combining an image with a prompt which requires the knowledge stored in the model's parameters to infer culturally specific meaning. The figure shows an image provided to various language models \cite{anthropic_claude_2024, agarwal_many-shot_2024, team_gemini_2024, openai_gpt-4_2024} models. The image alludes to ``The Importance of Being Earnest" by Oscar Wilde, a culturally significant play in Britain. Given the prompt ``Looking at the attached image to which 19th-century comedy play does it refer? and why?" the models generally returned the correct answer, a typical response and explanation from \cite{anthropic_claude_2024} is as follows\newline

\textit{This image appears to reference Oscar Wilde's ``The Importance of Being Earnest" (1895). The emoji equation shows a baby + handbag = shocked face, which refers to the play's plot point about a baby being accidentally left in a handbag at Victoria Station. This is a crucial element of the play's story, where it's revealed that one of the main characters was found as a baby in a handbag at the railway station's cloakroom. This mix-up forms one of the play's most famous and absurd plot points, leading to the memorable line about being found in ``a handbag" delivered with shock and horror by Lady Bracknell.}
\newline

This response demonstrates the importance of inter-leaving text and images. It is this ability to understand context and cultural references that makes multi-modal language models particularly promising for historical document OCR. It is these abilities that this paper attempts to leverage to accurately extract the text from the images stored in the NCSE dataset given the context of the prompt and text itself.

\subsection{Objectives and contributions of the paper}

Whilst the performance of the large frontier models will consistently outperform the small ones, they can be prohibitively expensive for a relatively small gain. This paper uses Pixtral 12b \cite{agrawal_pixtral_2024}, which balances cost with performance, to demonstrate the value of pre-trained image-to-text models as a cost-effective OCR tool for historical documents. It then applies the model to the NCSE \cite{brake_nineteenth-century_2008} collection in a pipeline to create a new, cleaned, classified and searchable dataset of 19th-century newspapers for the benefit of historians and social scientists. It demonstrates the value of the process by providing an initial analysis of the newly created dataset. Specifically, the paper seeks to answer the following core questions.

\begin{itemize}
    \item How can images be pre-processed to improve the performance of Pixtral as a system of OCR?
    \item Is an image-to-text model effective at performing OCR on historical documents?
\end{itemize}

In addition, it demonstrates the value of the resultant dataset, the NCSE v2.0, by answering the following questions.

\begin{itemize}
    \item What were the topic differences between the newspapers/periodicals?
    \item Were there any differences in the readability of the text?
    \item Over a comparable time-period how does the magnitude of coverage of a topic vary between periodicals?
\end{itemize}

Beyond the demonstration of image-to-text LMs as efficient OCR tools, the key output of this paper is the NCSE v2.0 dataset. This dataset is an update of the original NCSE dataset (which can be considered V1.0), with significantly improved OCR quality and bounding box locations, in addition the entries are classified into types of text and topic.

Note throughout this work the words ``newspaper" and ``periodical" will be used interchangably.

\FloatBarrier
\section{Data}

This paper has two different kinds of data. The data used to test the process and the data used to construct the NCSE v2.0 dataset.

\subsection{Test data}

This paper will use two datasets to test the quality of the OCR. The BLN600 \cite{booth_bln600_2024} and the NCSE test dataset. The BLN600 is a collection of 600 19th-century articles primarily related to crime. The NCSE test set is an improved version of the data used in \cite{bourne_clocr-c_2024, bourne_scrambled_2024-1}. The NCSE test set comprises 31 randomly selected pages from the NCSE dataset. The bounding boxes were manually drawn and classed using the Labelbox platform \cite{labelbox_labelbox_2025}. The boxes were then transcribed with line breaks such that the text is line aligned. The images are cropped to make using the dataset easier so that each bounding box comes in an image-text pair. This resulted in 358 image-text pairs.

\subsection{Data for NCSE v2.0}

The entire NCSE dataset is 4263 pdf files containing a total of 84,509 page images\footnote{This is about 12k lower than the NCSE metadata; this difference is due to multiple editions of the same paper being present in the metadata but not the image data.}.
As shown in Table \ref{tab:periodicals_table}, the dataset is quite diverse, with the periodicals covering a range of different years, significant differences in the number of pages,  and the physical size of those pages between periodicals. The Northern Star is over twice as high and twice as wide as the smallest periodical, The English Woman's Journal. These size differences impact processing due to the relative size of the text compared to the page. However, the physical size of the letters between periodicals is likely relatively similar as historically, the font size of print has fallen in a relatively narrow range \cite{legge_does_2011}. In terms of storage, the total size of the dataset is 60 GB. To make these image files easier to process, this paper takes advantage of the fact that the images are all black and white and converts them into 2-bit PNG files at 120 dots per inch (except the Northern star which is much physically larger so 200dpi is used). Converting the images in this way reduces the size of the total dataset to approximately 27Gb, a saving of 55\%. In addition, separating each page into a single file allows independent processing.

\begin{table*}
\caption{The newspapers and periodicals of the NCSE and their key information}
\label{tab:periodicals_table}
\begin{tabular}{p{4cm}lrrp{1cm}p{1cm}}
\toprule
Title & Years & Issues & Pages  & Width (cm)  & Height (cm)\\
\midrule
Monthly Repository and Unitarian Chronicle & 1806-1837 & 486 & 17,734  & 33 & 48 \\
Northern Star & 1837-1852 & 2226 & 17,965  & 51 & 78\\
Leader & 1850-1860 & 1005 & 24,391 & 36 & 49\\
English Woman’s Journal & 1858-1864 & 91 & 5,663 & 24 & 34 \\
Tomahawk & 1867-1870 & 188 & 1,996 & 35 & 38\\
Publishers’ Circular & 1880-1890 & 285 & 16,760 & 34 & 51 \\
\bottomrule
\end{tabular}
\end{table*}

\FloatBarrier
\section{Method}

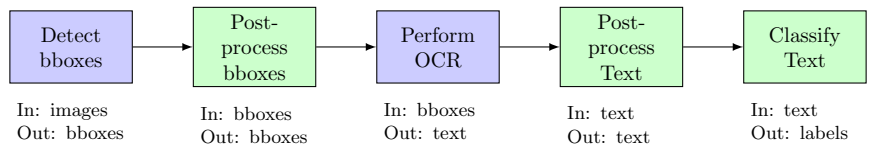
\begin{figure*}[htbp]  
    \centering    
\begin{tikzpicture}[scale=0.8, transform shape, 
    box/.style={
        draw,
        rectangle,
        minimum width=2cm,    
        minimum height=1.2cm,  
        align=center,
        text width=1.8cm,
        fill=blue!20  
    },
    redbox/.style={  
        draw,
        rectangle,
        minimum width=2cm,    
        minimum height=1.2cm,  
        align=center,
        text width=1.8cm,
        fill=green!20
    },
    io/.style={
        align=left,
        text width=1.8cm,     
        font=\small
    },
    >=latex
]

\node[box] (detect) {Detect bboxes};
\node[io, below=0.2cm of detect] (detect_io) {In: images\\ Out: bboxes};

\node[redbox, right=1cm of detect] (postbox) {Post-process bboxes};
\node[io, below=0.2cm of postbox] (postbox_io) {In: bboxes\\ Out: bboxes};

\node[box, right=1cm of postbox] (ocr) {Perform OCR};
\node[io, below=0.2cm of ocr] (ocr_io) {In: bboxes\\ Out: text};

\node[redbox, right=1cm of ocr] (posttext) {Post-process Text};
\node[io, below=0.2cm of posttext] (posttext_io) {In: text\\ Out: text};

\node[redbox, right=1cm of posttext] (classify) {Classify Text};
\node[io, below=0.2cm of classify] (classify_io) {In: text\\ Out: labels};

\draw[->] (detect) -- (postbox);
\draw[->] (postbox) -- (ocr);
\draw[->] (ocr) -- (posttext);
\draw[->] (posttext) -- (classify);

\end{tikzpicture}
    \caption{Overall project process showing inputs and outputs at each stage. Blue boxes use the image data, green boxes only require text, or structured data}
    \label{fig:pipeline}   
\end{figure*}

This section is broken into four parts: Layout detection and bounding boxes, OCR Extraction, Text classification, and Analysis of the NCSE v2.0. Figure \ref{fig:pipeline} shows the overall processing pipeline. Each stage of the pipeline has clear inputs and outputs, making substitution and comparison easy. In addition, it allows the performance at each stage to be evaluated independently. Although the details of each stage and the evaluation metrics are introduced in the following subsections, an overview is provided in table \ref{tab:stage_overview}.

\begin{table*}
\centering
\begin{tabular}{llll}
\toprule
&\textbf{Phase} & \textbf{Test Metrics} & \textbf{Datasets} \\
\midrule
1 & Create Bounding Boxes & Coverage (\%), Overlap (\%) & NCSE \\
2 & Perform OCR & Character Error Rate & NCSE, BLN600 \\
4 & Article Classification & Topic F1, Class F1 & NCSE \\
\bottomrule
\end{tabular}
\caption{Overview of pipeline phases with corresponding evaluation metrics and datasets used for testing.}
\label{tab:stage_overview}
\end{table*}

The modular structure of the evaluation is important as without a very highly controlled test set where the position of all characters is known, measuring the interaction between evaluation metrics is not possible. For example, if a bounding box overlaps part of a title and part of a text, it is not possible to evaluate the overall OCR quality.

\FloatBarrier
\subsection{Layout Detection and bounding boxes}

In the case of archival newspapers, layout detection refers to identifying articles, images, headlines, and so on found on the printed page; these regions of interest are then defined by bounding boxes.

Whilst the original NCSE dataset contained bounding boxes, these were found to be of very poor quality, often missing significant sections of the page, overlapping with other bounding boxes, or crossing multiple columns of text. As such, it was decided that the bounding boxes would be re-made using a pre-trained layout detection model.

Typically, layout detection uses some form of deep learning, whether this be language model-based \cite{huang_layoutlmv3_2022, shehzadi_bridging_2023} or a vision model using a convolutional neural network (CNN). This paper considers only models using the CNN due to the combination of high performance and low cost. Specifically, the paper uses DocLayout-Yolo. DocLayout-Yolo is based on the popular Yolo architecture using Yolo-v10, a 15.4 million parameter model \footnote{This is very small compared to most language models; for reference, it is almost 1000 times smaller than the Pixtral language model, used for the OCR} trained on 300k synthetic images. In addition, DocLayout-Yolo achieves State-Of-The-Art across several benchmarks.

Two alternative models were qualitatively evaluated; these were the Yolov8 model used by \cite{dell_american_2024} and also The RT-DET model used in the Docling pipeline, which was trained on 2k and 81k images, respectively. However, a qualitative analysis of the performance showed that the Docling model produced a large number of erratic bounding boxes, whilst the model by \cite{dell_american_2024} appeared to consistently miss areas of the page and having sizing issues; this may be due to it's relatively small training set.

One of the disadvantages of the DocLayout-Yolo model is that it is trained to work with a wide range of modern document types, not archival newsprint; this means that it can struggle with dense text-heavy images and also can classify the bounding boxes with impossible classes, such as ``equation". The model could be fine-tuned. However, sufficient training data is not available, and the data used by \cite{dell_american_2024} was not made publicly available. As such, the bounding boxes will be post-processed to increase the quality of archival newsprint.

\subsubsection{Post-processing bounding-boxes}
\label{sect:post_process_bbox}
The post-processing of bounding boxes is designed to minimise errors related to erroneous class (such as marking ``title" as ``equation"), poor alignment with columns, overlaps between bounding boxes, and gaps within columns. The post-processing does not use the images themselves; instead, it takes advantage of the inherent structure of the newspaper layout. The process utilizes the fact that text and images are rectangular, bounding boxes are exclusive and should not overlap, columns are generally equally sized, and the printed material covers as much of the page as possible.

The process of post-processing the bounding boxes is shown in Algorithm \ref{alg:bbox_post_process} (for detail, see the code for the function `postprocess\_bbox' in the `bbox\_functions' module). Overall, the process is as follows. The DocLayout-Yolo model has a bounding box class ``abandoned" which typically is page, numbers and periodical titles at the top and bottom of the page; this conveniently provides a way of getting rid of unnecessary boxes and miss-classified boxes in that region of the page. Next, the number of text columns on the page are found, and text boxes are assigned to the correct column number. The system is flexible enough to hand changes on a page; invalid box classes in the rest of the page, such as ``equation" or ``table caption" are often titles and so are mass re-classed as titles. The reading order of the bounding boxes is then found and used to adjust the bounding box's lower edge to the upper edge of the subsequent box This adjustment has the advantage of removing many of the overlapping boxes. Box widths are then extended such that boxes that do not cross the width of the column are extended to column limits, which prevents errors where lines are only partially included. Then, optionally, voids before the first bounding box in the column and after the last bounding box are filled with text bounding boxes; this is to ensure that columns are correctly filled. Then, for efficiency reasons, boxes less than a minimum height are removed, and smaller boxes are merged for efficiency when sending to the LM. The lower limits of boxes are re-adjusted, and the final reading order is calculated. Note for detail on any of the sub-processes in Algorithm \ref{alg:bbox_post_process} please refer to the ``bbox\_functions" module of the ``function\_modules" library in the codebase \cite{bourne_codebase_2025}.

\begin{algorithm}
\caption{Bounding Box Processing Algorithm}
\begin{algorithmic}[1]
\Require{\textit{bbox dataframe}, \textit{minimum height threshold}, \textit{fill columns}}
\State Reclassify boxes which will be abandoned
\State Remove abandoned boxes
\State Assign columns
\State Re-classify invalid classes as `title'
\State Create reading order
\State Adjust bbox lower limit to fill voids
\State Extend bbox x-limits to fill column
\If{\textit{fill columns}}
    \State Fill columns with bounding boxes of class `text'
\EndIf
\State Remove boxes with height $<$\textit{minimum height threshold}
\State Merge small boxes
\State Re-adjust bbox lower limit
\State Re-calculate reading order
\end{algorithmic}
\label{alg:bbox_post_process}
\end{algorithm}

Three different approaches will be taken. The first will be the output of DocLayout-Yolo, which is minimally processed to convert all the non-valid (``equation", etc.) classes to 'title'. The second approach will be the process shown in Algorithm \ref{alg:bbox_post_process}, without column filling, whilst the third will be with column filling. 

\begin{figure*}
    \centering
    \includegraphics[width=\linewidth]{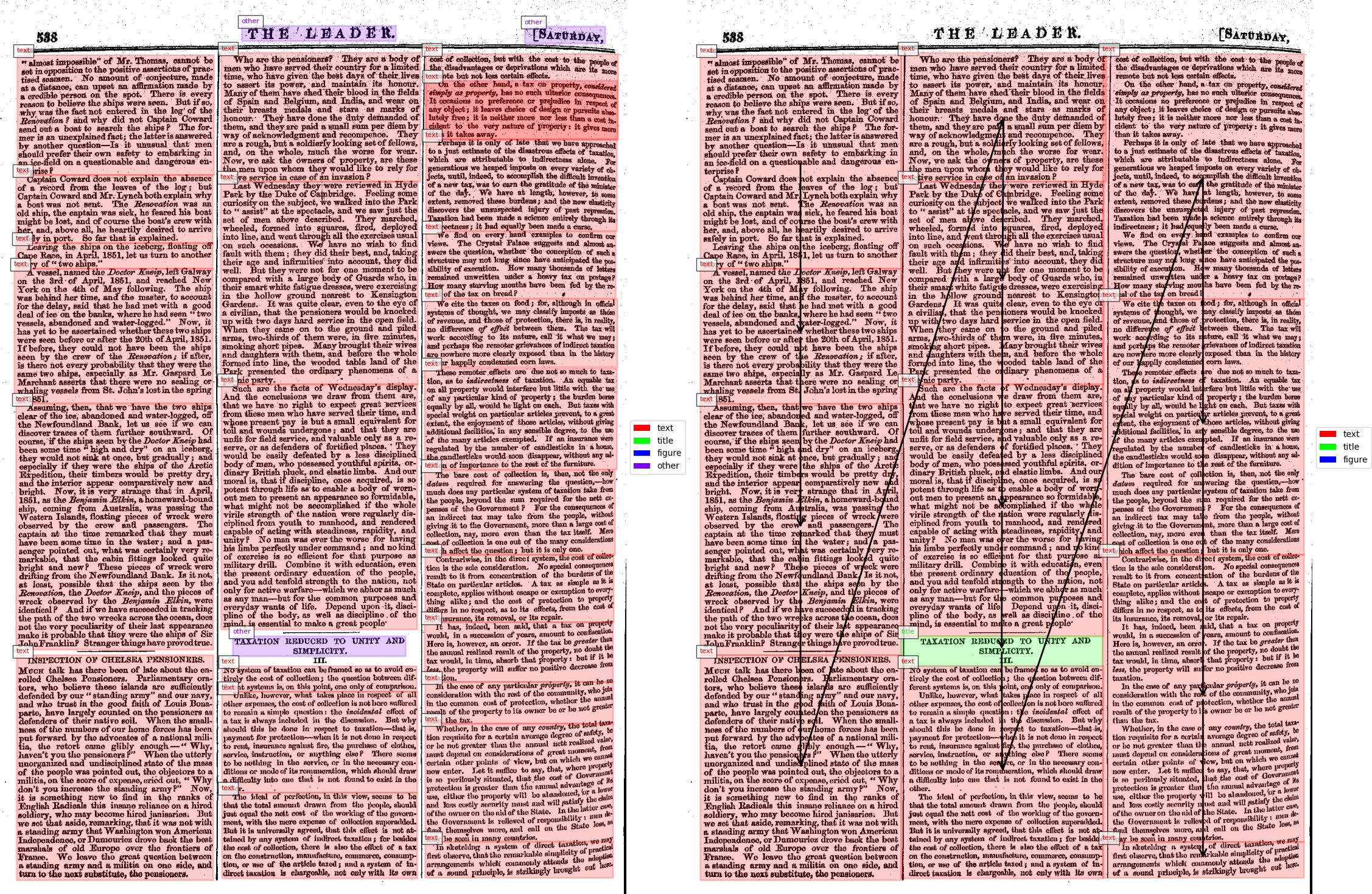}
    \caption{The affect of post-processing bounding boxes is shown for page `CLD-1852-06-05\_page\_14'. Post-processing caused box re-classification and the removal of overlaps as well as the addition of the reading order shown as black arrows. }
    \label{fig:CLD-1852-06-05_page_14_combine}
\end{figure*}

An example of the effect of post-processing bounding boxes is shown in Figure \ref{fig:CLD-1852-06-05_page_14_combine} for page `CLD-1852-06-05\_page\_14'. The title box was mis-classed when the original bounding boxes were created (blue in original, green post-processing). In addition, there was an overlap in the top right corner (seen as an area of darker red), which was fixed by the post-processing function. The blue boxes at the top of the page were correctly classed as abandoned by DocLayout-Yolo and were removed during post-processing. In addition, the reading order is shown as black arrows across the page; this is essential for correctly re-assembling the text post-OCR.

While the process can handle a diverse range of layouts, it also struggles to deal with issues such as asymmetrical columns, large overlaps in bounding boxes, and unusual combinations of bounding box class.

\subsubsection{Evaluating post-processing of bounding-boxes}

Evaluating the quality of bounding boxes and any post-processing without a custom test set is challenging. This is partly because the `correctness' of bounding boxes is subjective relative to the philosophy of how page design is viewed and the training approach of the object detection algorithm. Due to these challenges, the quality of the bounding boxes will be evaluated using the percentage overlap of all bounding boxes and the fraction of the printed area covered by bounding boxes. These metrics are simple and won't work well in some instances, for example, empty or nearly empty pages,  when a single bounding box is erroneously very large and covers all other correct boxes. However, without a detailed test set, they provide a solution that does not require a test set. In addition, these metrics have the flexibility needed to handle the changes to the bounding-box approach which the post-processing will produce. 

Each page's Coverage and Overlap scores are simply calculated using the expressions shown in equations \ref{eq:image_values} to \ref{eq:overlap}. First, a mask is created the same size as the image

\begin{equation}
    M(i,j) = \sum_{k=1}^n I\left[  p(i,j) \in b_k \right]
    \label{eq:image_values}
\end{equation}

Where for an image of height $h$ and width $w$ pixels, $B = \left\{  b_1, b_2, \cdots, b_n \right\}$ is the set of $n$ bounding boxes on the page, $p(i,j)$ is the pixel at position $i,j$ and $I\left[ \cdot \right]$ is the indicator function which is 1 if pixel $i$,$j$ is in bounding box $k$. This results in an image mask where the value of each pixel is the total number of bounding boxes the pixel is covered by.

Using this mask the Coverage is calculated using
\begin{equation}
    \text{Coverage} = \frac{\sum_{i=1}^h \sum_{j=1}^w (M(i,j) > 0)}{wh}
    \label{eq:coverage}
\end{equation}
such that the value is the fraction of all pixels who are covered by at least one bounding box.
Similarly Overlap is calculated as
\begin{equation}
    \text{Overlap} = \frac{ \sum_{i=1}^h \sum_{j=1}^w  (M(i,j) > 1)}{wh}
    \label{eq:overlap}
\end{equation}

which is the fraction of all pixels covered by at least two bounding boxes. As the values are all relative to the number of pixels in the original image, both Overlap and Coverage are bounded between 0 and 1.

It should be noted that these two metrics do not indicate that the bounding boxes are correct or appropriate. For example, a single bounding box that covered the entire print area of the page shown in Figure \ref{fig:CLD-1852-06-05_page_14_combine}, would appear perfect.

\subsection{OCR Extraction}

With the bounding box post-processing complete, it is time to consider the OCR approach itself. The process can be broken into three stages: image pre-processing, OCR evaluation and text post-processing.

As this paper is using an autoregressive LLM that has not been fine-tuned, an appropriate prompt needs to be given. Exploring prompt strategy is beyond the scope of this paper, and the prompt is simply designed to cover each of the three main categories `text' (including `title'), `figure' and `table'. The prompts used are shown in Table \ref{tab:prompts}. Although three categories of data are being extracted, this paper only focuses on the text class.

\begin{table*}[htbp]
\centering
\begin{tabular}{|p{0.15\textwidth}|p{0.75\textwidth}|}
\hline
\textbf{Prompt Type} & \textbf{Prompt Text} \\
\hline
Text & The text in the image is from a 19th century English newspaper, please transcribe the text including linebreaks. Do not use markdown use plain text only. Do not add any commentary. \\
\hline
Figure & Please describe the graphic taken from a 19th century English newspaper. Do not add additional commentary \\
\hline
Table & Please extract the table from the image taken from a 19th century English newspaper as a tab separated values (tsv) text file. Do not add any commentary \\
\hline
\end{tabular}
\caption{Prompts used for converting the data from the different image types into text form.}
\label{tab:prompts}
\end{table*}

\subsubsection{Pre-processing images for the LM}

\begin{figure}
    \centering
    \includegraphics[width=\linewidth]{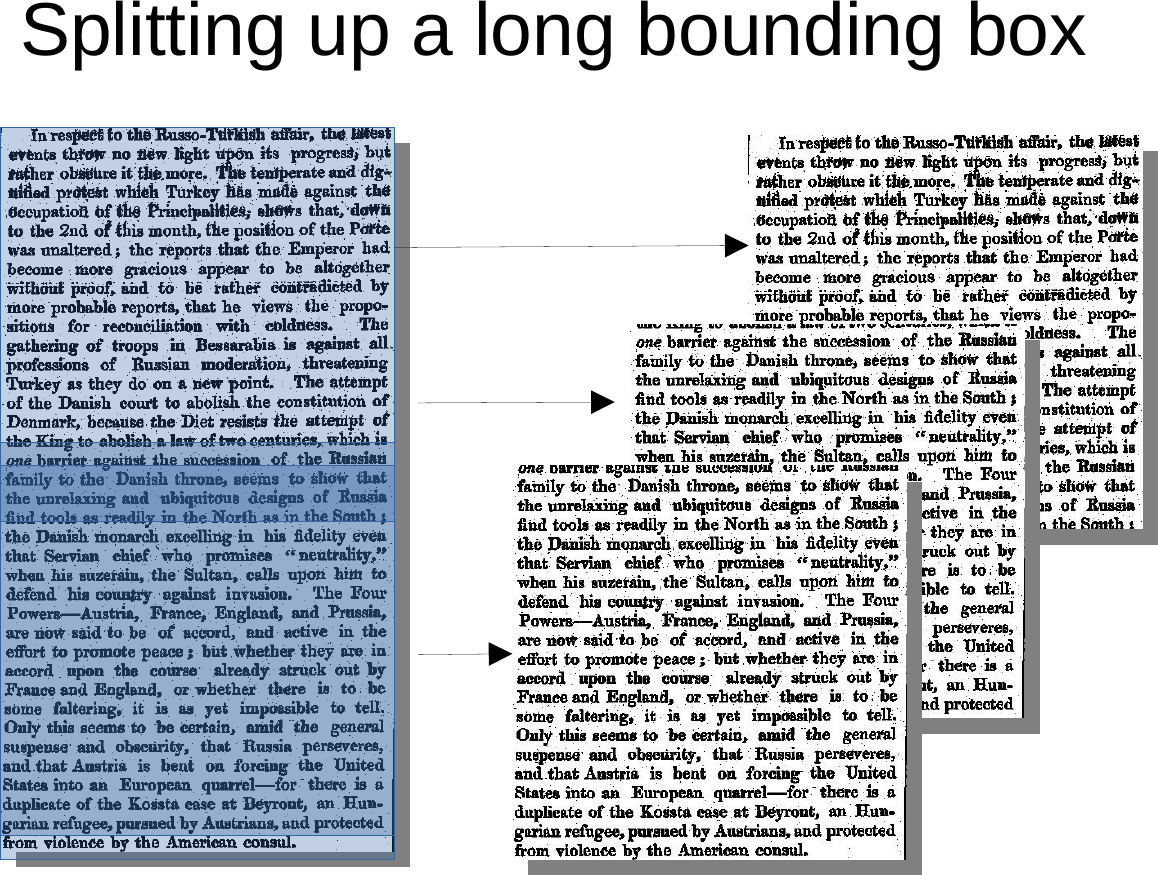}
    \caption{Bounding box `CLD-1853-07-30\_page\_2\_B0C2R7' is split into three boxes. Overlap is shown as darker shades of blue.}
    \label{fig:splitting_bounding}
\end{figure}

Pixtral uses a custom 16x16 vision encoder \cite{dosovitskiy_image_2021} to represent images \cite{agrawal_pixtral_2024}. The model is designed to handle rectangular images of any aspect ratio by flexibly dividing the image into 16 equally sized patches. They also employ 2D relative rotary position encodings \cite{su_roformer_2023} instead of absolute embeddings to handle changes in the aspect ratio and image size. This innovation is valuable for processing newspaper content because articles are rectangular but have no fixed aspect ratio. In addition, for fixed dots per inch, the size of the image may vary depending on the physical size of the paper. However, care needs to be taken because if the aspect ratio becomes too extreme, the patch embeddings may become quite long or wide relative to individual characters, affecting the ability of the model to understand the text. To check whether the aspect ratio impacts quality, OCR will be performed on the test sets by cropping the article image using aspect ratios of 1 (square crop), 1.5, 2, Inf (no crop). The piece-wise images will have a 20\% overlap to allow the full text to be reconstructed. If long bounding boxes negatively impact the OCR process, the cropped images will have a lower CER. Figure \ref{fig:splitting_bounding} shows how bounding box `CLD-1853-07-30\_page\_2\_B0C2R7' is split up when a square crop is used with 20\% overlap. The algorithm increases the area of overlap in the last box to ensure it is square. As the third box is only a single line, boxes two and three overlap almost completely. The overlap can be seen in the left image as areas of darker blue.

Whilst cropping the image can handle potential deformation issues within the representation created by the Language Model itself, the proportions of the image are not the only potential source of transcription errors. Rotated pages and warping caused by the image not lying completely flat can also cause difficulties in OCR \cite{girdhar_digitizing_2024, brown_document_2001}. A second experiment will be performed to automatically correct for page rotation using ``deskewing". As such, four pre-processing tests will be performed using all combinations of image cropping (True, False) and deskewing (True, False).

\subsubsection{Evaluation of OCR quality}
OCR quality will be compared using the BLN600 and NCSE test datasets. The NCSE test dataset will be cropped to article or paragraph level and paired with the ground truth text.

Four different OCR approaches will be compared: Pixtral (`pixtral-12b-2409'), Tesseract \cite{smith_overview_2007} (pytesseract 0.3.13), Efficient OCR \cite{carlson_efficient_2024}(3.2.2), EasyOCR (1.7.2), and GOT (0.1.0). Tesseract, first made open-source in 2005, has become the defacto OCR open-source software that can be considered a baseline level of quality. EasyOCR is newer but has become very popular for its ease of use and ability to extract text from photographs of natural scenes. Efficient OCR is the model used in \cite{dell_american_2024} and is the only model designed specifically for archival texts. GOT is a small image-to-text language model developed specifically for OCR purposes and the only other image-to-text model being evaluated. To directly compare the impact of model size, Pixtral-Large ('pixtral-large-2411') will also be used. This model is conceptually the same as Pixtral 12B but 124B parameters, meaning it should perform substantially better. However, Pixtral-Large will not be used to perform OCR on the entire dataset as, at the time of writing, it was 13x more expensive for input tokens and 40x more expensive for output tokens.

Each model will process all images, and the resultant text will be scored against the ground truth using the Character Error Rate (CER).

Character Error Rate (CER) is a commonly used metric for analysing the quality of text recovery in OCR tasks. 

\begin{equation}
    \textrm{CER} = \frac{S + D + I}{S + D + C}
\end{equation}

CER is defined as the sum of the character substitutions (S), deletions (D), and insertions (I), divided by the substitutions, deletions, and correct characters (C). Although not particularly sophisticated, CER provides a simple, intuitive and broadly reliable measure of text quality. It should be noted that CER can be greater than 1. This can happen if there is a substantial number of insertions, which can be the case even when the original text is otherwise recovered perfectly. Such occurrences can occur when the LM hallucinates writing text not present in the original. Using the median when measuring the CER across a range of documents is important, as a major hallucination event in a single document can significantly skew the final result when aggregating with the mean. The impact of using mean vs median can be seen in the contrasting results of \cite{boros_post-correction_2024} and \cite{bourne_clocr-c_2024} and is directly tested and discussed in \cite{bourne_clocr-c_2024}.

This paper is concerned with the accurate transmission of the meaning of the text; as such, line breaks, capitalisation, and extra spacing will be ignored when calculating the CER. 

\subsubsection{Post-processing of OCR results}

Two types of post-processing were performed on the dataset. The first is light touch stripping any beginning and end symbols such as `` ` "; the text of class `title' that is longer than 50 characters will be re-classed as text, and for consistency, line breaks will be removed such that a paragraph is a single line. After this, the dataset can be considered complete.

The second type of post-processing continues after the first, is more aggressive in its changes, and results in a dataset organised into article groups. This post-processing will take advantage of a quirk of 19th-century media, before the arrival of ``New Journalism," which was that there was minimal text formatting \cite{hampton_new_2008, twycross_rise_2024} and titles were predominantly all upper-case and paragraph separated. As such, text boxes will be split into paragraphs, and each paragraph will be tested as to whether it is a title or not using Algorithm \ref{alg:is_title}. The Algorithm ensures that the entire ``Paragraph'' in upper case is at least 5 letters and contains at least two vowels; in the case these three tests are True, the class will be changed to ``title"; otherwise, it will be kept as ``text". This approach allows for cases where the text contains a title at the beginning, middle or end or for cases where there are multiple titles within a text string.
After separating these embedded titles and re-classing, consecutive titles will be merged into a single title block. Articles will then be constructed by merging title rows with all subsequent text, figure, and table rows until the next title. This will result in a dataset where each row is, in theory, a single article. 

\begin{algorithm}
\caption{Is Title}
\label{alg:is_title}
\begin{algorithmic}[1]
\Require{text string $s$}
\Ensure{boolean indicating if string meets title criteria}

\Function{IsTitle}{$s$}
    \If{$s \neq \text{toUpper}(s)$}
        \State \Return false
    \EndIf
    
    \State $lettersOnly \gets \text{removeNonLetters}(s)$ \Comment{Remove non-letters and spaces}
    
    \If{$\text{length}(lettersOnly) < 5$}
        \State \Return false
    \EndIf
    
    \State $vowelCount \gets \text{countMatches}(s, \text{[AEIOU]})$
    
    \If{$vowelCount < 2$}
        \State \Return false
    \EndIf
    
    \State \Return true
\EndFunction
\end{algorithmic}
\end{algorithm}

The main limitation of this post-processing stage is that creating a ground truth is challenging as the amount of post-processing required is a result of the bounding-box method used and any bounding-box post-processing applied. As such, the existing ground truth cannot be used. Because of this, the title separation will only be evaluated qualitatively.

\subsection{Classification}

There will be two classification models created for the data. Classifying the type of text, classifying and the topic of the text using the Tier 1 IPTC subject codes \cite{iptc_news_2019}. The classes of these groups will be as follows.

\begin{itemize}
    \item Text Type: article; advert; poem/song/story; other
    \item Topic: arts, culture, entertainment and media; crime, law and justice; disaster accident and emergency incident; economy, business and finance; education; environment; health; human interest; labour; lifestyle and leisure; politics; religion; science and technology; society; sport; conflict, war and peace; weather
\end{itemize}

This creates, four text-type classes and 17 topic classes. 
      
To create the text-type model, 2000 samples of class 'text' will be selected from each newspaper to create a training set of 12,000 articles. This sampling approach ensures that each newspaper is equally represented and that there are sufficient data points to give examples of each class. Mistral Large (mistral-large-2411) will classify the texts into ``Text-Type". A ModernBert Model \cite{warner_smarter_2024} will then be trained and used to classify the entire dataset. Next, another 12,000 examples of class text or title (minimum of 30 tokens) will be selected; this dataset will be classified into one of the IPTC topics using Mistral Large. Like the Text-type classifier, a ModernBert Model will be trained and used to predict across the dataset. 

The training data will be split 80/20 for train and validation, and the models will be evaluated against the LLM classified dataset using the F1 score, shown in equations \ref{eq:precision}, \ref{eq:recall}, and \ref{eq:f1}. The analysis will use both macro and micro F1 to understand overall model performance, where micro F1 is the F1 for all observations in the data, whilst macro F1 is the mean F1 for individual classes.
For the single class classifier used to determine the exclusive text type classes, micro F1 is the same as accuracy.

\begin{equation}
\text{precision} = \frac{\text{True Positives}}{\text{True Positives} + \text{False Positives}}
\label{eq:precision}
\end{equation}

\begin{equation}
\text{recall} = \frac{\text{True Positives}}{\text{True Positives} + \text{False Negatives}}
\label{eq:recall}
\end{equation}

\begin{equation}
F_1 = 2 \cdot \frac{\text{precision} \cdot \text{recall}}{\text{precision} + \text{recall}}
\label{eq:f1}
\end{equation}

The ModernBert model is used for several reasons; it is very cost-efficient, having only 149M parameters, architectural changes relative to the original BERT family of models \cite{devlin_bert_2019, liu_roberta_2019, he_deberta_2021} have substantially improved both speed and performance. It should be noted that research has shown that traditional machine-learning approaches outperform auto-regressive language models like Mistral Large \cite{bohacek_when_2024}. However, the labelling approach will be considered acceptable when creating a training set. The ModernBert model will provide a consistent and cost-effective method to classify the entire dataset.

The basic training details are given in Table \ref{tab:bert_params}. All training and inference was done on an NVIDIA L4 with 24GB of ram. To reduce training time texts were truncated to the first 512 tokens. During inference batch size was increased substantially to 256. For more details on training and inference see the python scripts text\_type\_trainer.py, IPTC\_trainer.py and infer\_folder.py available at \cite{bourne_codebase_2025}.

\begin{table}
\centering
\begin{tabular}{ll}
\hline
\textbf{Parameter} & \textbf{Value} \\
\hline
Model Parameters & 149M \\
Learning Rate & 2e-5 \\
Batch Size & 32 \\
Maximum Tokens & 512 \\
Epochs & 5 \\
Warmup Ratio & 0.1 \\
Optimizer & AdamW \\
GPU & 24GB NVIDIA L4 \\
\hline
\end{tabular}
\caption{Training Parameters for the ModernBERT classifiers}
\label{tab:bert_params}
\end{table}

\subsection{Analysis}

The analysis of the NCSE v2.0 dataset will be simply a demonstration of potential use cases. A selection of 2000 texts from the class ``article" will be randomly selected from each newspaper to create a subset of 12000 observations. The distribution of topic labels will be compared, and the similarity between the newspapers in terms of the distribution will be measured using cosine similarity and hierarchical clustering \cite{mullner_modern_2011}. In addition, the Flesch-Kincaid readability \cite{flesch_new_1948} will be calculated using the bootstrapped median of the sub-set \cite{efron_introduction_1993}. Finally, The Leader and Northern Star have a period of overlap covering the ``Great Exhibition" of 1851 \cite{luckhurst_great_1951}. The Great Exhibition was a world fair of culture and technology making it news worthy in itself. However, it was also housed in the ``Crystal Palace" which was at time the largest glass building in the world \cite{addis_crystal_2006}. As such there was likely considerable discussion of the exhibition and the Crystal Palace at the time. For this analysis the count of database entries, which include the phrases "Crystal Palace" or "Great Exhibition", will be plotted by month for both newspapers.

\section{Results}

The results section is divided into five subsections, covering different aspects of the OCR and dataset creation process. 

\subsection{Post-processing bounding-boxes}

By measuring the overlap and coverage of the bounding boxes produced by the two post-processing methods relative to the bounding boxes produced by DocLayout-Yolo, we can evaluate which post-processing, if any, should be chosen. The results show that the two post-processing have a similar level of performance controlling for periodical. Both methods generally reduce overlap or leave it unchanged and increase coverage. However, in the case of the Northern Star, the column-filling approach provides an increase of almost 10 percentage points over the more straightforward post-processing method and 24 percentage points compared to no post-processing. This increase takes the total median coverage to 100\% compared to 92\% and 79\% for simple post-processing and no-post processing, respectively. However, this increased coverage comes at the cost of an increase in mean overlap from 17\% to 24\%.
The Northern Star behaves differently from the other periodicals because it has much larger pages which contain more columns.

As a result, all the papers will use simple post-processing without column filling, apart from the Northern Star, which will use column filling.

\subsection{Image pre-processing}

\begin{figure*}
    \centering
    \includegraphics[width=\linewidth]{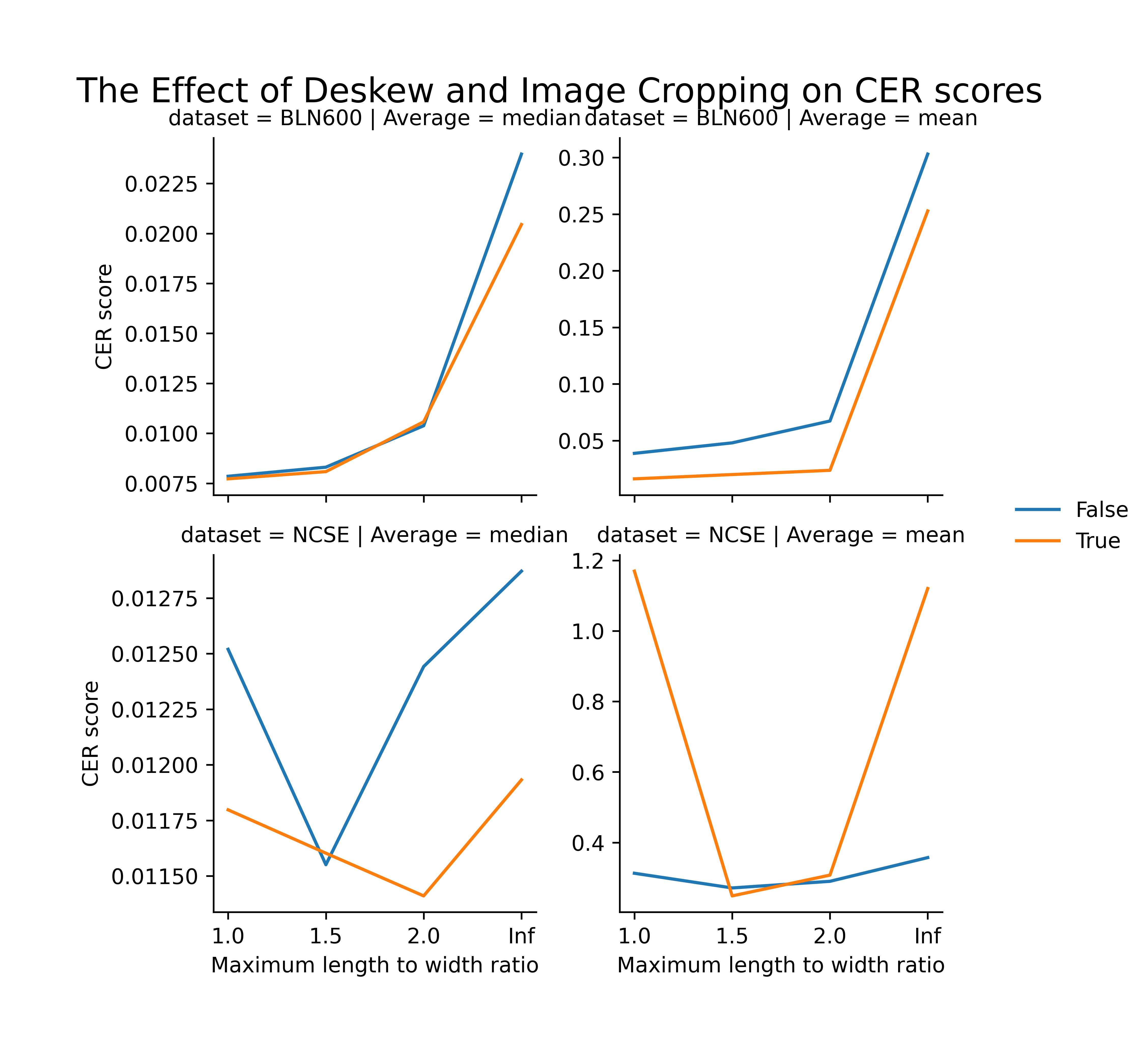}
    \caption{Pre-processing the image can significantly decrease the CER of the returned text. There is a much larger decrease in the mean CER than the median, indicating that the probability of major errors in a text is substantially reduced by pre-processing}
    \label{fig:enter-label}
\end{figure*}

Figure \ref{fig:enter-label} shows the result of the experiments in pre-processing the region of interest before sending to the LM. The results show that without cropping, the mean CER is much larger than the median due to substantial OCR errors. These errors occur for both datasets and with and without deskewing. The errors tend to be repeating a token or groups of tokens until the token limit is reached or simply commenting that the image looks like a newspaper text. Such responses can get CER values much higher than one, even into the hundreds. The impact of deskew is more nuanced, as it causes a consistent and significant reduction in the mean CER for BLN600, but the effect on the mean NCSE CER is unclear. In addition, the median CER is only marginally affected. Interestingly, the impact that a larger image window has on the total number of tokens is very different between the two datasets. BLN600 has a mean total token of 2740 for no crop and 8950 for a square crop, an increase of 320\%. In contrast, the NCSE test set has 2210 for no crop and 2635 for square crop, an increase of only 20\%. This difference is likely due to the ratio of the bounding boxes themselves; BLN600 uses the entire image as a bounding box, whilst NCSE is more like paragraph level. As a result of this analysis, this paper will not use deskew; the image crop ratio will be 1.5, as will the merge-height in the bounding-box post-processing discussed in section \ref{sect:post_process_bbox}.

\subsection{Comparison of OCR quality}

Table \ref{tab:model_results} compares Pixtral, Tesseract, EasyOCR, and Efficient OCR and GOT for both the NCSE test set and BLN600. The table shows that Pixtral comprehensively outperforms all other methods. In addition, the other methods failed to beat the Tesseract baseline.

Whilst Table \ref{tab:model_results} shows the median results, the mean results were also calculated and compared. All models had lower median scores than the mean, except for Efficient OCR, which scored better on the mean NCSE test set. The difference between Mean and Median performance was particularly noticeable for Pixtral and GOT on the NCSE dataset. Pixtral got a mean of 0.27, and GOT got a mean of 1.25. These errors were primarily driven by getting locked into repeating a set of characters until the token limit was reached. Pixtral Large was also used to extract the text from the datasets to test the impact of parameter size on performance. Both Pixtral models have the same median CER. However, the mean is quite different. Pixtral large has a mean CER score of only 0.01 and 0.06 on the BLN600 and NCSE test set, respectively, much lower than Pixtral 12B. This substantial difference is because unlike Pixtral 12B or GOT, Pixtral Large does not get stuck repeating phrases. 

However, the overall difference between the two Pixtral models is slight; of the 958 observations across both datasets, 98\% Pixtral-Large CER scores were less than 0.1; in contrast, Pixtral 12B achieved this 96\% of the time, showing just how skewed the distribution is and the importance of using the median for evaluation.

As well as having the best overall performance, Pixtral was the model most likely to have zero errors, with 7\% of the responses having a CER of 0. GOT had 1.5\% with zero error, and Tesseract had 0.5\%. All results from EasyOCR and EffOCR had some errors.

\begin{table}
\caption{The results show Pixtral outperforms all other models}
\label{tab:model_results}
\begin{tabular}{lll}
\toprule
Model & BLN & NCSE \\
\midrule
GOT & 0.41 & 0.06 \\
EasyOCR & 0.28 & 0.26 \\
EffOCR & 0.35 & 0.62 \\
Pixtral & \textbf{0.01} & \textbf{0.01} \\
Tesseract & 0.05 & 0.05 \\
\bottomrule
\end{tabular}
\end{table}

\subsubsection{Performance on example text}

\begin{figure}
    \centering
    \includegraphics[width=\linewidth]{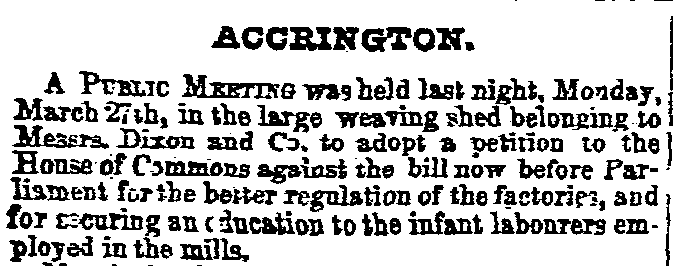}
    \caption{An excerpt from the Northern Star 1st April 1843, with densely packed text and poor quality printing. This produced broadly average results across the models}
    \label{fig:image_example}
\end{figure}

Figure \ref{fig:image_example} shows an image taken from Northern Star 1st April 1843,  \footnote{For reference to the data set, the bounding box ID is `NS2\_1843-04-01\_page\_4\_B0C1R1'}. The print is not particularly clear, with some smudging and streaking evident. To provide a more intuitive example of the errors, the first 200 characters of the returned texts are shown in Table \ref{tab:NS_example}, along with the CER for the entire text. The models generally perform close to the test set median, except GOT which became stuck in a loop repeating `Honsc of Commons agmist the bill now before Par-' until the token limit was reached, producing a massive CER of 35. This kind of error is typical of, and unique to, the errors observed in LLM-based OCR, and both GOT and Pixtral suffer from it. If it had not been for this repeating error, GOT would have been a close second to Pixtral. The example also shows how critical low errors are to create a usable reference text; even error rates of 11\% produce a substantial amount of unreadable text.

\begin{table*}
\caption{The first 200 characters of the Northern Star Example in ascending order of error.}
\label{tab:NS_example}
\small
\begin{tabular}{p{1.2cm}p{0.8cm}p{9cm}}
\hline
Model & CER & Result \\
\hline
Pixtral & 0.01 & accrington.  a public meeting was held last night, monday, march 27th, in the large weaving shed belonging to messrs. dixon and co. to adopt a petition to the house of commons against the bill now bef \\
Tesseract & 0.11 & accrington.  a posuic mzerme was held last night, monday, march zish, in the large weaving shed belonging to! messra. dizon and c3, to adopt @ petition to the | honse of commons against the bill now b \\
EasyOCR & 0.28 & aggringtox,  4 prauc mxgroe pag beld jast pigbì mondaj , marcb 271b, in the feaving shed belonging i0 messra diron ad c5: adopt 8 petition lo the honse ol csmmors against the billnow before far hiamen \\
EffOCR & 0.58 & accblngtqn, 4 feb1ic meeiie8 wa held lat niphfmonday 4i4a8eu q *q, iii me 1afes weaving snfq petpdp1nr 10  442ihth4 9ne deter regulation of the factqf3f:. ! avi g@eql14\& an t3nea1ou to the i1fagt 1abd \\
GOT & 34.97 & accrngton. a peblc mrtng wacheld last night, monday, march 27th, in the large weaving shed belonging to messrs. dixon and co. to adopt a petition to the honsc of commons agmist the bill now before par \\
\hline
\end{tabular}
\end{table*}

\subsection{Creating the NCSE v2.0}

With the image pre-processing method decided and the clear result of the relative performance of the Pixtral model, the entire NCSE dataset was processed, creating the NCSE v2.0. Creating the dataset required 3.5 billion prompt tokens sent to the model and 599 million completion tokens generated, for a total token count of approximately 4 billion tokens. At the time of writing the cost of using the Pixtral api with batching (upto 24 hour delay but 50\% cheaper) was \$0.075 per million for both prompt and generation, this makes the cost of OCR for NCSE v2.0 about $4e^9 \times \$0.075e^{-6}= \$300$. The data set contains 82,690 pages of recovered data, which is 98\% of the original image pages, 321 million words and 1.9 billion characters. As would be expected given the variance shown in Table \ref{tab:periodicals_table}, there are large differences in the volume and type of information between periodicals; the details of this can be seen in Table \ref{tab:NCSE_v2}. The dataset has 1.1 million text entries, 198 thousand titles, descriptions of 17 thousand figures, and 16 thousand tables. After the more aggressive post-processing of the OCR, there were 841 thousand articles, which, given the total number of text boxes, seems high, suggesting the post-processing method is unreliable for improving title separation.

\begin{table*}
\caption{Key information on the NCSE v2.0 dataset}
\label{tab:NCSE_v2}
\begin{tabular}{p{4cm}lrrp{1cm}p{1cm}p{1cm}}
\toprule
Periodical & Issues & Pages & Figures & Tables & Text & Titles \\
\midrule
English Woman’s Journal & 91 & 5638 & 16 & 27 & 11766 & 1541 \\
Leader & 1005 & 23691 & 1539 & 8351 & 217207 & 43985 \\
Monthly Repository and Unitarian Chronicle & 329 & 17425 & 353 & 157 & 71571 & 7525 \\
Northern Star & 2215 & 17930 & 1907 & 3476 & 778923 & 120210 \\
Publishers’ Circular & 236 & 16174 & 12524 & 3180 & 60873 & 22097 \\
Tomahawk & 187 & 1832 & 959 & 534 & 7161 & 3000 \\
\bottomrule
\end{tabular}
\end{table*}

The ModernBert models performed relatively well. The text-type classifier had an accuracy/micro F1 of 0.85 across the four classes and a macro F1 of 0.78, reflecting the over-representation of the ``article" class. The IPTC-topic classifier had a similar level of performance; across the 17 classes, it had a micro F1 of 0.79 and a macro F1 of 0.72, reflecting the poor performance of some of the smaller classes. As a result, the models were used to classify the entire dataset by text type and topic.

\subsection{Analysis}

\begin{figure}
    \centering
    \includegraphics[width=\linewidth]{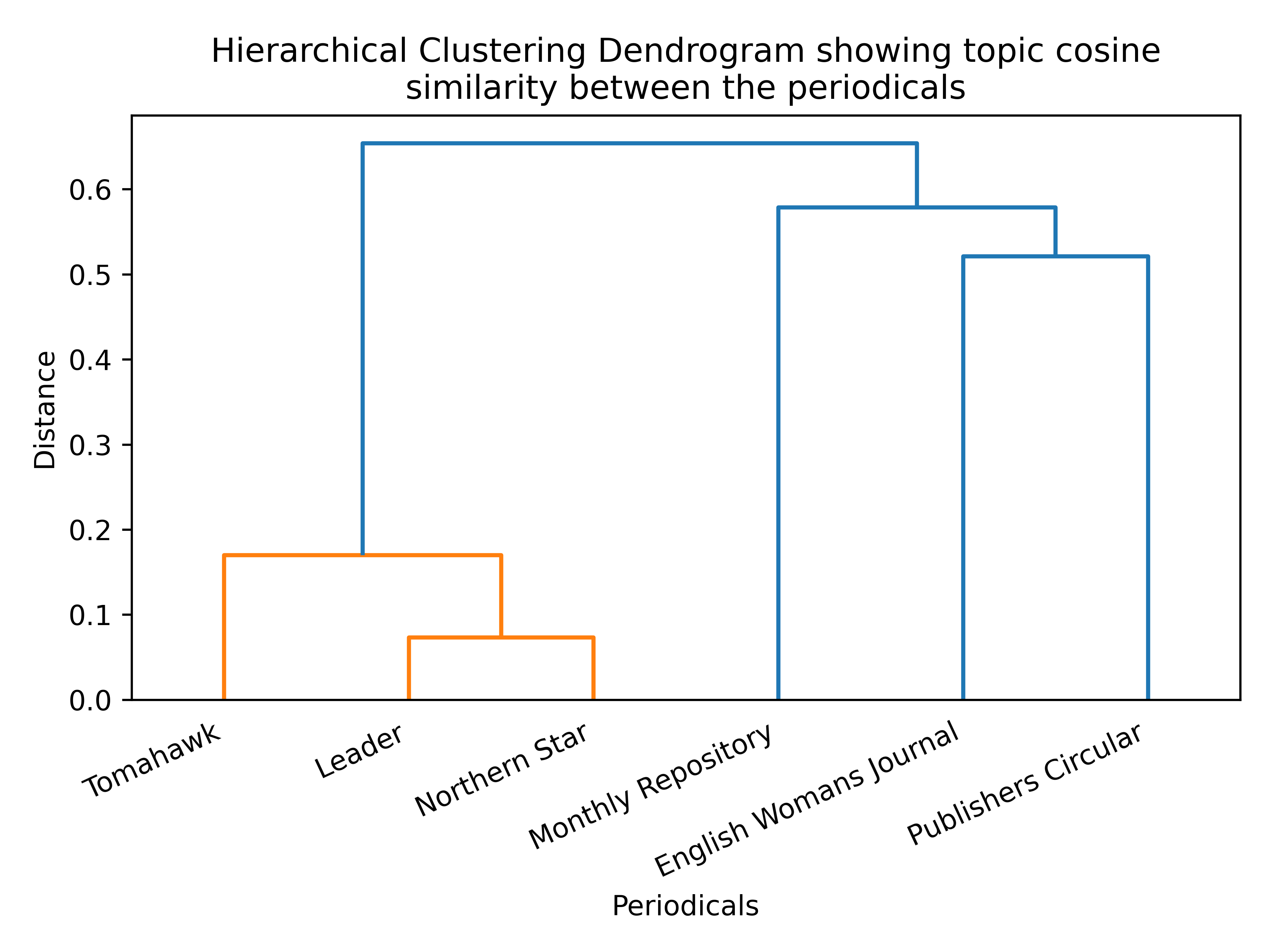}
    \caption{Periodical similarity based on topic distribution shows that the Leader, Northern Star, and Tomahawk are relatively similar due to the heavy focus on Politics; in contrast, the other three periodicals have quite distinct topic distributions. Distance is measure between 0 and 1.}
    \label{fig:topic_similarity}
\end{figure}

Table \ref{tab:top_cats}, shows the most frequently occurring topics by each periodical. There were substantial differences in the popularity of topics between the newspapers, which makes sense considering their focus. The English Woman's Journal's top topics of Human Interest, Society and Labour are unsurprising given that they were advocating for equal rights for women; likewise, Almost 50\% of the topics of the Monthly repository were related to Religion. The Leader had substantial coverage of Conflict War and Peace as it was active during the Crimean War \cite{badem_ottoman_2010, goldfrank_origins_2014} and the Second Opium War \cite{phillips_saving_2012}. Although there was a lot of variety in the most popular topic for each periodical, the least popular were broadly similar, with the Environment, Sport, and Weather seldom covered among any of the periodicals. The topic similarity of the periodicals was measured using cosine similarity visualised in Figure \ref{fig:topic_similarity} using hierarchical clustering. The Figure shows that the Northern Star and the Leader are very similar, followed closely by the Tomahawk; these three newspapers share a common interest in politics. In contrast, the English Woman's Journal, Publishers Circular, and Monthly Repository are quite distinct due to their focus on Woman's rights, Arts/Entertainment, and Religion, respectively.

\begin{table*}
\caption{Most common IPTC topic categories by periodical}
\label{tab:top_cats}
\begin{tabular}{ll}
\toprule
Periodicals & Top Categories \\
\midrule
English Womans Journal & Human Interest, Society, Labour \\
Leader & Politics, Conflict War Peace, Business/Finance \\
Monthly Repository & Religion, Politics, Human Interest \\
Northern Star & Politics, Crime Law Justice, Labour \\
Publishers Circular & Arts/Entertainment, Education, Business/Finance \\
Tomahawk & Politics, Human Interest, Arts/Entertainment \\
\bottomrule
\end{tabular}
\end{table*}

\begin{figure}
    \centering
    \includegraphics[width=\linewidth]{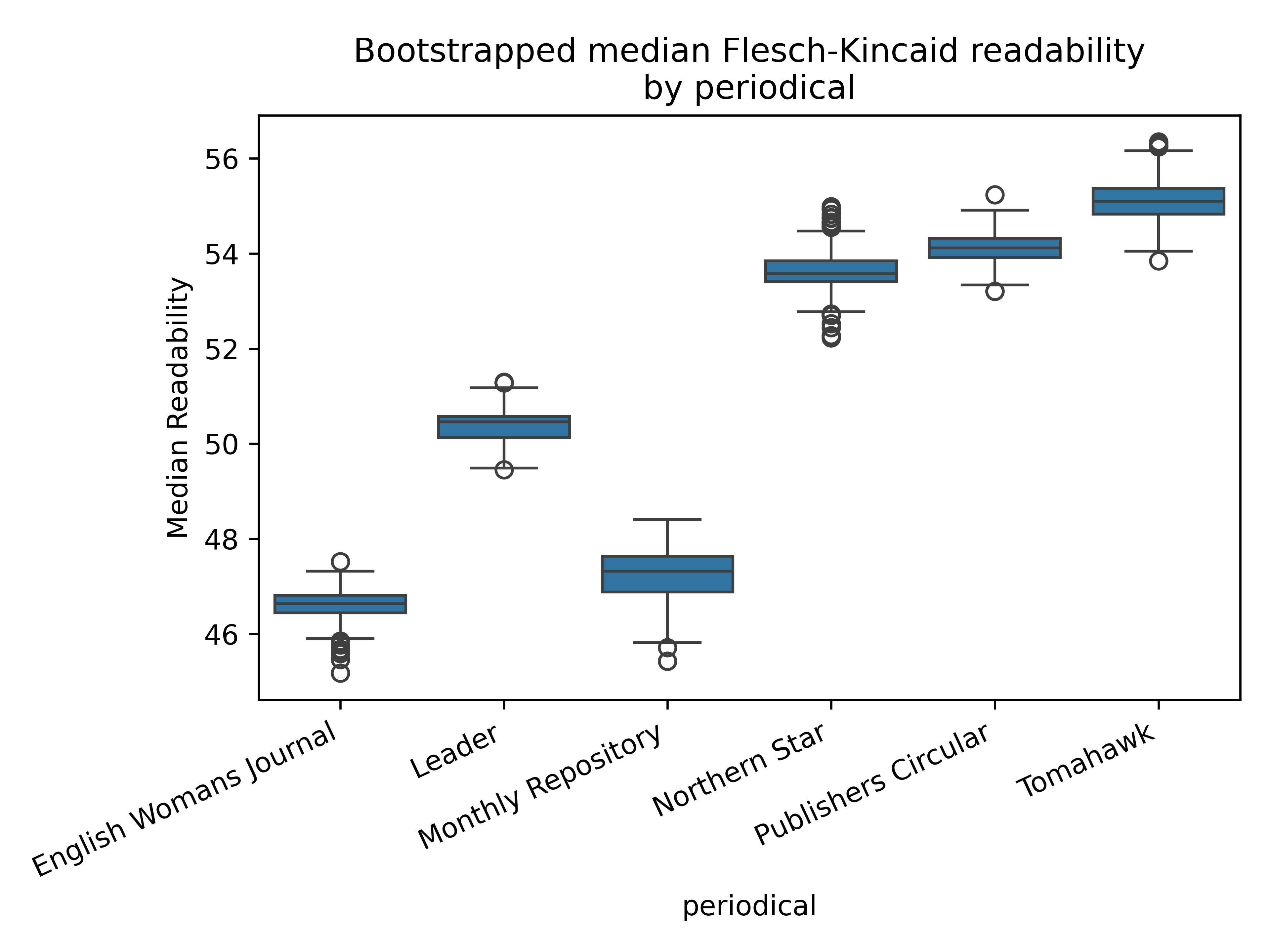}
    \caption{Although there are differences in the average readability, all the periodicals are broadly similar. By modern standards, they would be classed as relatively difficult to read.}
    \label{fig:readability}
\end{figure}

The median bootstrapped Flesch-Kincaid readability of the sampled periodicals was analysed as is shown in Figure \ref{fig:readability}. Despite their substantial topic differences and large temporal spread, they all had a similar reading level, roughly the range 45-55 out of 100, which by modern standards is considered fairly difficult to read. It should also be noted that the archaic nature of the newspapers may add some noise to the readability score. 

\begin{figure}
    \centering
    \includegraphics[width=\linewidth]{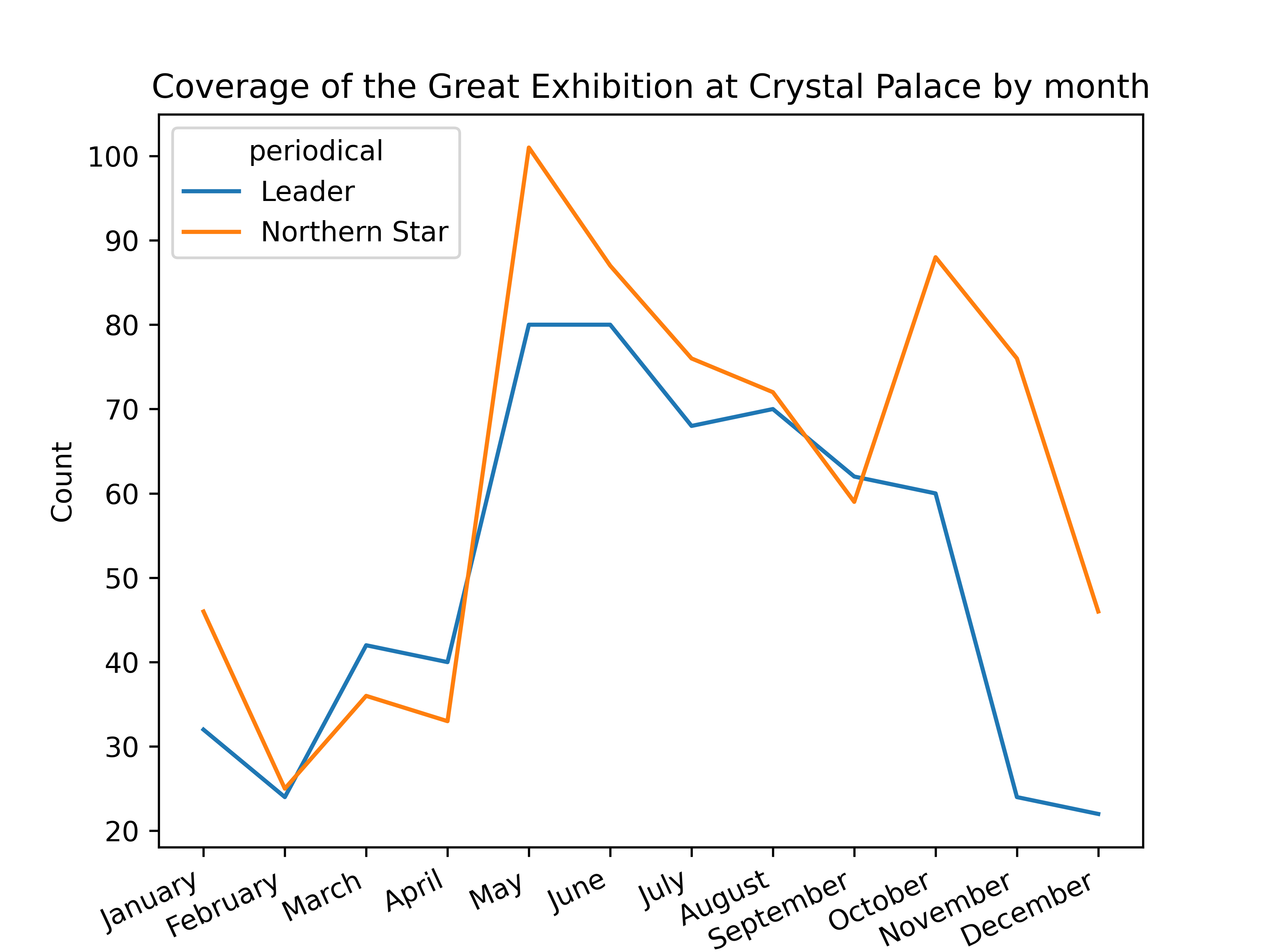}
    \caption{A comparison of the coverage of the 1851 Great Exhibition for the Leader and Northern star.}
    \label{fig:crystal_palace}
\end{figure}

Finally, Figure \ref{fig:crystal_palace} shows the Leader and Northern Star's coverage of the Great Exhibition in the Crystal Palace. The figure shows that coverage jumps in May when the exhibition started and drops in December as the exhibition closed in November. Although both Newspapers follow the same pattern, the Northern Star has more coverage overall, with 745 mentions to 604.

\FloatBarrier
\section{Discussion}

With only light pre-processing of images, an off-the-shelf image-to-text model dramatically outperformed four purpose-built OCR engines on historical document digitisation tasks. Pixtral obtained a median CER of 0.01 on both datasets, an error rate that can be described as "excellent" quality \cite{hodel_general_2021}. In addition, no model, apart from Pixtral, beat the Tesseract baseline, indicating that the current deep learning training sets and benchmarks are not appropriate for the noisy text-dense images common with historical documents. 

Despite being the only method explicitly designed for historical documents, Efficient OCR substantially underperformed its published results. This could be because, whilst Efficient OCR had a substantial silver training set, there was only a small hand-transcribed ground truth. In addition, the test set was only 275 lines of text. This may have led to over-fitting during training and/or an unrepresentative test set, leading to overconfidence in performance.

In many ways, this paper raises the same implicit critique of OCR that CoPali \cite{faysse_colpali_2024} does, which is that traditional OCR fundamentally misses the point when it comes to how text, particularly news media, is read and how it is designed to be read, which is as a whole. By combining the OCR abilities of the Pixtral LM with the page-parsing ability of DocLayout-Yolo, this paper creates a functional solution to the tension between character extraction and the greater semantic context. This combination results in very high-quality OCR within the bounding boxes. However, it lacks the page-level representation of CoPali, which can lead to missing areas of text and overlapping text boxes. However, this drawback is mitigated by raw text output, which is much more accessible and shareable.

The results in this paper show that the most serious issue relating to using language models with OCR is the risk of hallucination, particularly in getting locked into repeating phrases or characters. This phenomenon was experienced by both Pixtral and GOT but not by Pixtral Large. Cropping images substantially reduces the risk of major errors; however, fine-tuning would be the most effective approach without increasing model size. Fine-tuning has been shown to allow smaller models to compete with larger equivalents on the specific tasks they have been fine-tuned on. Therefore, it is reasonable to believe that fine-tuning Pixtral-12B could perform similarly to Pixtral-Large but at a much lower cost.

As mentioned earlier, an issue with fine-tuning is the lack of datasets. However, research has shown that LMs can be trained and fine-tuned on entirely synthetic data \cite{gunasekar_textbooks_2023, li_textbooks_2023, tan_15-pints_2024, wei_general_2024, bourne_scrambled_2024-1}. Synthetic historical documents could be generated by creating a system that converts text to a synthetic newspaper layout. Subsequently, corrupting the synthetic images with noise \cite{gupte_lights_2021} would allow the training of Language models robust to the issues observed in this paper. Such highly controlled data would solve many of the measuring issues this work has faced, such as estimating the quality of title identification, and promoting a straightforward way of measuring total image-to-text transcription quality, which was mentioned at the beginning of the method.

This paper also shows the limitation of using Language models for post-OCR correction \cite{soper_bart_2021, boros_post-correction_2024, thomas_leveraging_2024, bourne_clocr-c_2024, bourne_scrambled_2024-1, kanerva_ocr_2025}. In \textcite{bourne_clocr-c_2024}, the base OCR error was 0.17 and the best performance 0.07. However, it is now clear that the original NCSE dataset had very poor bounding boxes. In contrast, in this paper, with accurate bounding boxes, the base case Tesseract achieved a CER score of 0.05. As such, it can be seen that post-OCR correction is no substitute for a robust bounding-box process. Such a process would have high page coverage and low overlap, accurately bounding the elements of the page. Due to the lack of sufficient training data, the creation of synthetic datasets would also provide a substantial boost in training object detection models and possibly help avoid the errors observed when applying layout models trained on modern data \cite{zhao_doclayout-yolo_2024, auer_docling_2024}. However, consideration can also be given to the speed quality trade-off as CNN-type models such as YOLO are extremely fast but are not able to consider the page holistically in the same way as an image-to-text model such as \cite{faysse_colpali_2024}. The impact on process quality may be such that it is worth having more expensive bounding box detection to increase the page coverage and decrease the overlap of the bounding boxes, using some sort of image-to-text model.

Classifying the text and articles into text-type and IPTC-topic is valuable as a method of understanding the text. In this paper, the classification method was crude, with a silver set of data created using an LLM, then a much simpler classifier trained and used to predict across the entire dataset. The performance of the two classifiers was acceptable for the paper. However, the training dataset could be made more robust using data-centric cleaning methods \cite{northcutt_pervasive_2021}. Similar to how the IPTC classification system was used for the texts, adverts could be classified by sector using already existing taxonomies \cite{iab_interactiveadvertisingbureautaxonomies_2025}

Although brief, the analysis of the NCSE v2.0 gives an idea of possible use cases when historical documents are stored as text. It is easy to see that the study could be extended using other techniques such as sentiment analysis \cite{medhat_sentiment_2014, wankhade_survey_2022,}, vector embedding \cite{pilehvar_embeddings_2020, smits_fully-searchable_2025} and Retrieval Augmented Generation  \cite{lewis_retrieval-augmented_2020, hogan_large_2025}, or Discourse Analysis \cite{atkinson_qualitative_2000} to name a few.

\section{Conclusion}

This paper has demonstrated that pre-trained image-to-text models like Pixtral 12B are highly effective for performing OCR on archival documents. Pixtral significantly outperformed the other approaches, having a CER of 0.01 on both datasets, 5x lower than the second-best approach, Tesseract. This performance was despite not being designed specifically for OCR. The results demonstrate the flexibility and general-purpose nature of modern LMs. In addition to the OCR quality, the process only cost about \$3.7 per 1000 pages, making it appropriate for low-resource projects. 

The most significant limitations in the current process are poor bounding box quality and the Language Model getting stuck and repeating sections of text. Both of these issues could be substantially improved in future work using synthetic data and fine-tuning.

The process developed in this paper resulted in the creation of the NCSE v2.0 dataset. The dataset contains 84,000 pages of data across 1.4 million entries and 321 million words. All entries are categorized by type and topic. Analysis of the NCSE v2.0 showed that with high-quality OCR and clear classification, quantitive comparative analysis of archival documents is made substantially easier.

It is hoped that the method developed in this paper and the resultant dataset `NCSE v2.0' will reduce the difficulty of studying early news media because while the idea that ``The one duty we owe to history is to rewrite it." \cite{wilde_critic_1891} is certainly to be avoided in historical document OCR; everyone can agree that history should be both readable and read.

\section{Data Availability}

The Data used in this paper is available at the UCL data repository ``NCSE v2.0: A Dataset of OCR-Processed 19th Century English Newspapers" \cite{bourne_ncse_2025}. The repository contains the NCSE test set, the NCSE v2.0 at the bounding box level, the bounding-box dataset, and the IPTC and text-type silver sets, all data is made available under a Creative Commons Attribution 4.0 International License (CC BY).  All code for the project is available at the github repository for the project \cite{bourne_codebase_2025}.

The original NCSE page images can be found at the relevant King's College London Repositories \cite{turner_ncse_2024, turner_ncse_2024-1, turner_ncse_2024-2, turner_ncse_2024-3, turner_ncse_2024-4, turner_ncse_2024-5}, or linked to at the central project archive \cite{turner_nineteenth-century_2024}.

\printbibliography 

\end{document}